\documentclass[conference]{IEEEtran}
\IEEEoverridecommandlockouts
\usepackage{cite}
\usepackage{amsmath,amssymb,amsfonts}
\usepackage{algorithmic}
\usepackage{graphicx}
\usepackage{textcomp}
\usepackage{xcolor}
\usepackage{hyperref}
\def\BibTeX{{\rm B\kern-.05em{\sc i\kern-.025em b}\kern-.08em
    T\kern-.1667em\lower.7ex\hbox{E}\kern-.125emX}}
\begin{document}

\title{Systematic analysis of the impact of label noise correction on ML Fairness
\thanks{This work was partly funded by: Agenda ''Center for Responsible AI'', nr. C645008882-00000055, investment project nr. 62, financed by the Recovery and Resilience Plan (PRR) and by European Union -  NextGeneration EU.; AISym4Med (101095387) supported by Horizon Europe Cluster 1: Health, ConnectedHealth (n.o -- 46858), supported by Competitiveness and Internationalisation Operational Programme (POCI) and Lisbon Regional Operational Programme (LISBOA 2020), under the PORTUGAL 2020 Partnership Agreement, through the European Regional Development Fund (ERDF); and Base Funding - UIDB/00027/2020 of the Artificial Intelligence and Computer Science Laboratory - LIACC - funded by national funds through the FCT/MCTES (PIDDAC).}}

\author{
    \IEEEauthorblockN{Inês Oliveira e Silva}
    \IEEEauthorblockA{\textit{Faculdade de Engenharia da Universidade do Porto} \\
    Porto, Portugal \\
    up201806385@edu.fe.up.pt}
    \and
    \IEEEauthorblockN{Carlos Soares}
    \IEEEauthorblockA{\textit{Faculdade de Engenharia da Universidade do Porto} \\
    \textit{Laboratory for Artificial Intelligence and Computer Science (LIACC)}\\
    \textit{Fraunhofer Portugal AICOS} \\
    Porto, Portugal \\
    csoares@fe.up.pt}
    \and
    \IEEEauthorblockN{Rayid Ghani}
    \IEEEauthorblockA{\textit{Machine Learning Department}, \\
    \textit{Heinz College of Information Systems
    and Public Policy}, \\ 
    \textit{Carnegie Mellon University}\\
    Pittsburg, PA, USA \\
    ghani@andrew.cmu.ed}
    \and
    \IEEEauthorblockN{Inês Sousa}
    \IEEEauthorblockA{\textit{Fraunhofer Portugal AICOS} \\
    Porto, Portugal \\
    ines.sousa@fraunhofer.pt}
}

\maketitle
\begin{abstract}
Arbitrary, inconsistent, or faulty decision-making raises serious concerns, and preventing unfair models is an increasingly important challenge in Machine Learning. Data often reflect past discriminatory behavior, and models trained on such data may reflect bias on sensitive attributes, such as gender, race, or age. One approach to developing fair models is to preprocess the training data to remove the underlying biases while preserving the relevant information, for example, by correcting biased labels. While multiple label noise correction methods are available, the information about their behavior in identifying discrimination is very limited. In this work, we develop an empirical methodology to systematically evaluate the effectiveness of label noise correction techniques in ensuring the fairness of models trained on biased datasets. Our methodology involves manipulating the amount of label noise and can be used with fairness benchmarks but also with standard ML datasets. We apply the methodology to analyze six label noise correction methods according to several fairness metrics on standard OpenML datasets. Our results suggest that the Hybrid Label Noise Correction~\cite{xu2020hybrid} method achieves the best trade-off between predictive performance and fairness. Clustering-Based Correction~\cite{nicholson2015label} can reduce discrimination the most, however, at the cost of lower predictive performance.
\end{abstract}


\section{Introduction}
\label{sec:introduction}

The widespread use of ML systems in sensitive environments has a profound impact on people's lives when given the power to make life-changing decisions~\cite{mehrabi2021survey}. One well-known example is the Correctional Offender Management Profiling for Alternative Sanctions (COMPAS) software. This computer program assesses the recidivism risk of individuals and is used by the American courts to decide whether a person should be released from prison. In a 2016 investigation conducted by ProPublica~\footnote{\href{https://www.propublica.org/article/machine-bias-risk-assessments-in-criminal-sentencing}{https://www.propublica.org/article/machine-bias-risk-assessments-in-criminal-sentencing}}, it was discovered that the system was biased against African-Americans, incorrectly classifying Black offenders as ``high-risk'' twice as often as White defendants. Another example relates to a less impactful yet more widely present tool in people's lives: Google's targeted ads. A group of researchers proposed AdFisher~\cite{datta2014automated}, a tool to gather insights on how user behaviors, Google's transparency tool ``Ad Settings'', and the presented advertisements interact. Their study revealed that male web users were more likely to be presented with ads for high-paying jobs than their female counterparts. In this context, we can classify an algorithm as unfair if its decisions reflect some kind of prejudice or favoritism towards certain groups of people based on their inherent or acquired characteristics~\cite{mehrabi2021survey}.

The process of learning which factors are relevant to the desired outcome in these tasks involves generalizing from historical examples, which can lead to algorithms being vulnerable to the same biases that people projected in their past decisions. For example, Amazon's ML experts tried to build a recruiting engine to automate the review of job applicants' resumes. However, they realized that their tool was discriminatory towards women. This was suspected to be the consequence of using training data from the previous ten years, during which most technical positions were applied for and granted to men, leading the system to discard most female applicants\footnote{\href{https://www.reuters.com/article/us-amazon-com-jobs-automation-insight/amazon-scraps-secret-ai-recruiting-tool-that-showed-bias-against-women-idUSKCN1MK08G}{https://www.reuters.com/article/us-amazon-com-jobs-automation-insight/amazon-scraps-secret-ai-recruiting-tool-that-showed-bias-against-women-idUSKCN1MK08G}}.

The goal of \emph{fair machine learning} is to identify and mitigate these harmful and unacceptable inequalities~\cite{barocas-hardt-narayanan}. When collecting data, prejudice will lead to incorrect labels, as the relationship between an instance's features and its class will be biased. Despite the vast amount of literature on methods for dealing with noisy data, only a few of these studies focus on identifying and correcting noisy labels~\cite{nicholson2015label}. This approach of correcting wrongly attributed labels can be leveraged in the context of fair machine learning if we consider discrimination present in the data as noise that can be removed. As such, noise correction techniques can be applied to obtain a feasibly unbiased dataset that can be used to train fair models. Thus, the motivation for this work comes from, to the best of our knowledge, the lack of work exploring the use of label noise correction techniques in training fair models from biased data.

We develop an empirical methodology to systematically evaluate the usefulness of applying label noise correction techniques to guarantee the fairness of predictions made by models trained on biased data. Having an assumedly clean dataset, we first manipulate the labels to simulate the desired amount and type of label noise. The injected noise is group-dependent, meaning that it depends on the value of the specified sensitive attribute. We can parameterize the noise injection process to model various types of discrimination. The considered label noise correction technique is applied to the noisy data to generate a corrected version of the dataset. We train ML classifiers using the \textit{original}, \textit{noisy}, and \textit{corrected} training sets. The obtained models are then evaluated under different assumptions, measuring the fairness and predictive performance of their predictions on the three test sets (\textit{original}, \textit{noisy}, and \textit{corrected}).

In this empirical study, we test and compare the effectiveness of six label noise correction techniques in improving the generated models' performance. We apply our methodology using multiple standard ML datasets available on OpenML and inject different types of label noise at varying rates. The models are evaluated using four well-known fairness metrics.  

The rest of this paper is organized as follows. In Section~\ref{sec:relatedwork}, we present an overview of the existing literature and state-of-the-art methods related to ML fairness and label noise. In Section~\ref{sec:methodology}, we propose the methodology to systematically evaluate the impact of label noise correction methods on the fairness of ML models. In Section~\ref{sec:experiments}, we describe the performed experiments and analyze the obtained results in Section~\ref{sec:results}, presenting the corresponding discussion in Section~\ref{sec:discussion}. Finally, in Section~\ref{sec:conclusions}, we review the conclusions that were derived from the developed work.

\section{Related work}
\label{sec:relatedwork}

In this section, we introduce the relevant literature related to label noise and dealing with fairness under label noise.

\subsection{Label noise}

Noise can be defined as non-systematic errors that might complicate an algorithm's ability to uncover the relationship between the features and the class label of a sample~\cite{frenay2013classification}. When noise is related to wrongly assigned labels, we are in the presence of label noise. Label noise is a common phenomenon in real-world datasets, and the cost of acquiring non-polluted data is usually high. This makes it of great importance to develop methods that deal with this type of noise~\cite{algan2021image}.

Label noise is particularly important in the case of bias mitigation techniques, which typically assume the existence of clean labels. However, in practice, this is not always the case. In fact, data bias and label corruption are closely related, especially since the accuracy of certain labels is often affected by the subject belonging to a protected group~\cite{wang2021fair}. Label noise can be classified into one of three categories: 
\begin{itemize}
    \item \textbf{Random noise}, which corresponds to noise that is randomly distributed and does not depend on the instance's features or label~\cite{algan2021image}, i.e., $P(\tilde{y}) = P(y)$;
    
    \item \textbf{Y-dependant noise} happens when instances belonging to a particular class are more likely to be mislabeled~\cite{algan2021image}. This type of label noise assumes that given the clean label $y$, the noisy label $\tilde{y}$ is conditionally independent of the instance $x$, i.e., $P(\tilde{y}|y,x) = P( \tilde{y} | y )$~\cite{wu2022fair};
    
    \item \textbf{XY-dependant noise} depends on both features and target values, meaning that the probability of a sample being mislabeled changes not only according to its particular class but also to the values of its features~\cite{algan2021image}. This is the type of noise commonly referred to as group-dependant~\cite{wang2021fair} or instance-dependant~\cite{wu2022fair} in the fairness literature. This type of label noise is often related to discrimination. Considering the COMPAS case, for example, the model unfairly predicts African-Americans as having a ``high risk'' of recidivism more often than Caucasians due to discrimination in past trials, which leads to models that reproduce the same kind of discrimination. In this situation, the probability of an offender being misclassified as ``high risk'', i.e., the label noise, depends on the \textit{race} feature, so it is group-dependant.

\end{itemize}

One way to categorize noise-dealing approaches is to classify the existing methods according to whether they model the noise structure or not~\cite{algan2021image}. Noise model-free methodologies focus on algorithms that are inherently less sensitive to label noise and thus do not require the explicit modeling of the noise structure. On the other hand, the goal of noise model-based methods is to extract information about the noise structure in the data to leverage it during training~\cite{algan2021image}. The label noise correction methods we focus on in this work and further present are included in this category.

\subsection{Fairness in the presence of Label Noise}

While many methods have been proposed to promote the fairness of ML classifiers, these usually assume that the training data is not corrupted~\cite{wang2021fair}. However, label noise is a common phenomenon in real-world data that may have negative consequences on model performance when not properly dealt with~\cite{frenay2013classification}. 

One approach to achieve fair classification is to focus on re-weighting the training data to alter its distribution in a way that corrects for the noise process that causes the bias~\cite{jiang2020identifying}. The authors have shown that training on the re-weighted dataset is equivalent to training on the unobserved unbiased labels. To evaluate how their method performed in comparison to previous approaches, they tested the various methods on a number of benchmark fairness datasets, measuring multiple fairness metrics.

A different line of work focuses on enforcing fairness constraints on the learning process to achieve fair predictions. Research has been conducted in adapting this approach for learning fair classifiers in the presence of label noise~\cite{wang2021fair, wu2022fair}. Some authors rewrite the loss function and fairness constraints to deal with label noise~\cite{wu2022fair}. They further propose to model label noise and fairness simultaneously by uncovering the internal causal structure of the data. Surrogate loss functions and surrogate constraints have also been devised to ensure fairness in the presence of label noise~\cite{wang2021fair}.


\subsection{Evaluation of Robustness}

The performance of label noise correction methods depends on the level of noise in the data. They are expected to improve fairness by correcting possible biases. For practitioners to apply those methods safely in the real world, it is important to understand their behavior under different noise conditions. However, there is currently a lack of research in understanding how those techniques affect the fairness of models. 

To address this limitation, a sensitivity analysis framework for fairness has been developed~\cite{fogliato2020fairness}. It assesses whether the conclusions about the fairness of a model derived from biased data are reliable. This is done by estimating bounds on the fairness metrics under assumptions about the magnitude of label noise. However, this approach still relies on a limited set of fairness benchmarks, limiting the scope of the conclusions since the existing datasets are not representative of many different types and levels of label noise. 

In this work, we address the limitations of the empirical evaluation procedures that are usually conducted in the existing work. Instead of making assumptions about the level of label noise, we explicitly manipulate it.

\section{Methodology}
\label{sec:methodology}

With the objective of understanding the effect of existing label correction methods on improving the fairness of machine learning classifiers trained on the corresponding corrected data, we propose a methodology for empirically evaluating the efficacy of such techniques in achieving this goal.

Having the \textit{original} dataset, $D_o$, in which we assume the instances to have correctly assigned labels, the first step is to manipulate the labels. 
When considering a fairness benchmark dataset, we may use the data as expected, meaning that the sensitive attributes and positive class are the original ones. If we are applying this methodology to a standard classification dataset, we arbitrarily choose the positive class and a binary attribute to be considered as the sensitive one.
Given noise rate $\tau$, noise injection is performed by altering the label of instances with a certain probability depending on the noise rate and whether it belongs to the protected group. By parameterizing this process, we can simulate different types of discrimination. We thus obtain a \textit{noisy} dataset, $D_n$, that is corrupted by the induced bias.

To simulate different types of biases, we inject group-dependant label noise in the clean datasets in two ways:

\begin{itemize}
    \item \textbf{Positive Bias Noise}. This type of label noise is intended to simulate the cases where the instances belonging to the protected group are more likely to be given a positive label (or the ones not belonging to the protected group are systematically assigned to the negative class). For example, this would be equivalent to classifying African-American offenders as having a high risk of re-offending at a higher rate than their Caucasian counterparts. To simulate such bias, we set the label of each instance belonging to the protected group to the positive class with a probability equal to the desired noise rate. Naturally, as the noise rate gets higher, the data gets progressively more imbalanced.

    \item \textbf{Balanced Bias Noise}. In different situations, both the members of the protected group are benefited and the non-members are harmed. An example of this type of scenario is the automated selection of job applicants. If the selection process is biased towards preferring male applicants, there will be simultaneously more men being selected and more women being rejected. We simulate such bias by setting the label of each instance to the positive class if it belongs in the protected group or to the negative class otherwise, with a probability equal to the desired noise rate. We assume that the positive class is a good outcome, which is not always the case. Nevertheless, this is not expected to affect the conclusions.

\end{itemize}

The next step is to perform label noise correction by applying the method being analyzed on the training set, obtaining a \textit{corrected} training set, $D_c$. We first examine the similarity between the original labels and the ones obtained after applying label noise correction to the noisy data. Given a dataset with $N$ instances, the ability to reconstruct the original labels is measured as the similarity between the \textit{original} labels and the \textit{corrected} ones, as shown in Eq.~\ref{eq:similarity}. Essentially, this is a measure of the accuracy of the label correction method in obtaining the original labels. However, to avoid confusion, we will refer to it as \emph{reconstruction score}, $r$.

\begin{equation}
    r = \frac{\sum_{i=1}^N \hat{y}_i = y_i}{N}
    \label{eq:similarity}
\end{equation}

For each training set ($D^{train}_o$, $D^{train}_n$, and $D^{train}_c$), we then apply the chosen ML algorithm to it, obtaining the classifiers $M_o$, $M_n$, and $M_c$, respectively. These models are then evaluated under different scenarios. 

Firstly, we want to consider the testing scenario where we only have access to corrupted data both for training and testing. The aim is to understand the effect of correcting training data in the case where the discrimination that was present when collecting the training data still exists at testing time. To achieve this, the \textit{corrected} ($M_c$) and \textit{noisy} ($M_n$) models are evaluated on the \textit{noisy} test set, $D^{test}_n$. In this case, the intent is to observe if the noise correction methods are able to produce less discriminatory predictions without significant loss in predictive performance.

Our next objective is to understand the effect of correcting biased training data when the discrimination has been eliminated in the meantime and the testing data is unbiased. To achieve this, the models ($M_o$, $M_n$, and $M_c$) are evaluated on the \textit{original} test set $D^{test}_o$.

Finally, we extend the previous scenario to remove the assumption that the original data is unbiased. In other words, we analyze the effect of correcting training data when the discrimination has been eliminated in the meantime but the original data was already biased and, thus, its labels are noisy. To achieve this, the \textit{corrected} model, $M_c$, is evaluated on a test set with labels without noise. However, since we do not have access to the clean labels, we use a label noise correction method to correct the test data as well. We employ the same method that is being analyzed, but a more extensive empirical validation could use different methods or a combination of them. In any case, the results should be interpreted carefully, as the unbiased labels cannot be determined.

\begin{figure}[ht]
    \centering
    \includegraphics[width=0.45\textwidth]{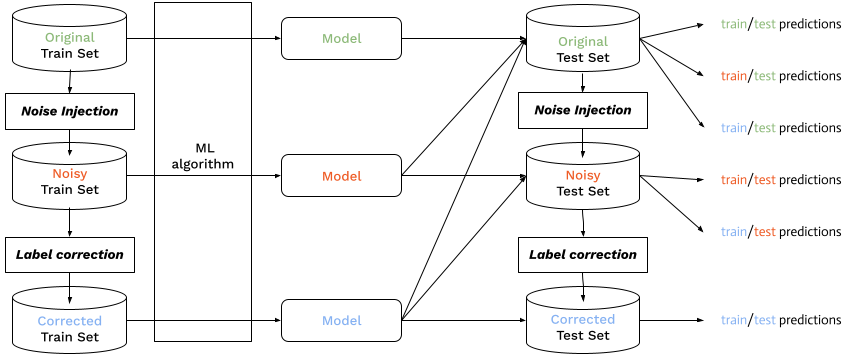}
    \caption{Diagram of the proposed methodology for empirically evaluating the efficacy of label noise correction methods in ensuring the fairness of classifiers using standard ML datasets.}
    \label{fig:methodology_injection}
\end{figure}

The diagram presented in Fig.~\ref{fig:methodology_injection} illustrates the explained methodology.

\section{Experiments}
\label{sec:experiments}

To illustrate the use of the proposed methodology, we perform an empirical evaluation of six label noise correction methods to ensure the fairness of ML models. In this section, we describe its key aspects, detailing how we evaluated the methods, explaining the experimental setup, and analyzing the obtained results.

\subsection{Label noise correction methods}

\begin{table*}[!ht]
    \centering
    \caption{Characterization of the datasets used in the conducted experiments.}
    \label{tab:datasets}
    \small
    \begin{tabular}{l|p{0.06\linewidth}llp{0.085\linewidth}p{0.09\linewidth}p{0.15\linewidth}p{0.17\linewidth}}
    \hline
        dataset & OpenML id & \# inst. & \# feat. & inst. in positive class & inst. in protected group & inst. in positive class and protected group & inst. in positive class and unprotected group \\ \hline 
        ads & 40978 & 1377 & 1558 & 33 \% & 76 \% & 34 \% & 33 \% \\ 
        bank & 1461 & 15111 & 30 & 33 \% & 51 \% & 24 \% & 43 \% \\ 
        biodeg & 1494 & 1055 & 41 & 34 \% & 15 \% & 5 \% & 39 \% \\ 
        churn & 40701 & 2121 & 22 & 33 \% & 23 \% & 21 \% & 37 \% \\ 
        credit & 29 & 653 & 43 & 45 \% & 31 \% & 47 \% & 45 \% \\ 
        monks1 & 333 & 556 & 6 & 50 \% & 49 \% & 49 \% & 51 \% \\ 
        phishing & 4534 & 11055 & 30 & 56 \% & 66 \% & 59 \% & 49 \% \\ 
        sick & 38 & 636 & 26 & 33 \% & 39 \% & 12 \% & 47 \% \\ 
        vote & 56 & 312 & 14 & 58 \% & 52 \% & 54 \% & 63 \% \\ \hline
    \end{tabular}
\end{table*}

We focus on problems where fairness issues are essentially caused by label noise, with the corruption rates being group-dependent. By assuming that there exist underlying, unknown, and unbiased labels that are overwritten by the observable biased ones, the natural approach is to apply label noise correction techniques to the data in order to obtain a clean dataset to be used in model training. The goal of this approach is to pre-process biased data to remove underlying discrimination, thus enabling classifiers trained on the corrected datasets to deliver predictions that are both accurate and fair.
In the conducted experiments, we compared the following label noise correction methods:

\begin{itemize}
    
    \item \textbf{Bayesian Entropy Noise Correction} (BE)~\cite{sun2007identifying}. In this method, multiple Bayesian classifiers are obtained with different training samples. These classifiers are used to obtain a probability distribution for each sample of it belonging to each considered class, which is applied in calculating the instance's information entropy. If the entropy of a sample is below the calculated threshold and its label is different from the predicted one, its value is corrected. These steps are repeated until a stopping criterion has been met;

    \item \textbf{Polishing Labels} (PL)~\cite{nicholson2015label}. This method replaces the label of each instance with the most frequent label predicted by a set of models obtained with different training samples;

    \item \textbf{Self-Training Correction} (STC)~\cite{nicholson2015label}. This algorithm works by first dividing the data into a noisy and a clean set using a noise-filtering algorithm. These methods identify and remove noisy instances from data, and in this case, the Classification Filter~\cite{triguero2014characterization} is used. A model is obtained from the clean set and is used to estimate the confidence that each instance in the noisy set is correctly labeled. The most likely mislabeled instance is relabeled to the class determined by the classifier and added to the clean set. These steps are repeated until the desired proportion of labels is corrected;
    
    \item \textbf{Clustering-Based Correction} (CC)~\cite{nicholson2015label}. Firstly, a clustering algorithm is executed on the data multiple times, varying the number of clusters. A set of weights is calculated for each cluster based on its distribution of labels and size and is attributed to all the instances that belong to it. These weights are meant to benefit the most frequent class in the cluster. The weights obtained from each clustering are added up for each instance, and the label with the maximum weight is chosen;

    \item \textbf{Ordering-based Label Noise Correction} (OBNC)~\cite{feng2015class}. The first step in this algorithm is to learn an ensemble classifier from the data. For each instance, the ensemble decides the label by voting, and the difference between the votes can be used to calculate an ensemble margin. The misclassified samples are ordered in a descending manner based on the absolute value of their margin. The most likely mislabeled instances are relabeled to their predicted classes;

    \item \textbf{Hybrid Label Noise Correction} (HLNC)~\cite{xu2020hybrid}. In this approach, the first step is to separate the data into high-confidence and low-confidence samples. This is achieved by applying the k-means algorithm to divide the data into clusters and determining each cluster's label. The instances are classified as high-confidence if their label matches the cluster's label and low-confidence otherwise. The high-confidence samples are used to simultaneously train two very different models, using the SSK-means~\cite{basu2002semi} and Co-training~\cite{blum1998combining} algorithms. These are applied to each low-confidence sample, and if both algorithms give it the same label, then the sample is relabeled and set as high-confidence. This process is repeated until all labels are high-confidence or after a specified number of times.

\end{itemize}


\subsection{Datasets and Algorithm}

The datasets used in the noise injection experiments are available on OpenML~\footnote{\href{https://www.openml.org/}{https://www.openml.org/}} and are summarized in Table~\ref{tab:datasets}.

In the conducted experiments, we used the Logistic Regression algorithm to obtain the classifiers. The code implementing the proposed methodology is available at \href{https://github.com/reluzita/fair-lnc-evaluation}{https://github.com/reluzita/fair-lnc-evaluation}.

\subsection{Evaluation Measures}

To evaluate the obtained models, we tested the predictive performance of the predictions by calculating the Area Under the ROC Curve (AUC) metric~\cite{hanley1982meaning}. In terms of fairness, the following metrics were analyzed:

\begin{itemize}
    \item \textbf{Demographic Parity} (also known as statistical parity) is a statistical group fairness notion that is achieved when individuals from both protected and unprotected groups are equally likely to be predicted as positive by the model~\cite{dwork2012fairness}. We analyze the Demographic Parity difference
    between the two groups:
    
    \begin{equation}
        DP_{dif} = |P(\hat{y}=1|g=0) - P(\hat{y}=1|g=1)|
    \end{equation}
    
    \item \textbf{Equalized Odds}~\cite{hardt2016equality} is satisfied when protected and unprotected groups have equal true positive rates (TPR) and equal false positive rates (FPR). To calculate the Equalized Odds difference, $EOD_{dif}$ between groups, we first obtain the TPR difference: 
    

    \begin{equation}
    \begin{split}
    TPR_{dif} & = |P(\hat{y}=1|y=1,g=0) \\&- 
    P(\hat{y}=1|y=1,g=1)|
    \end{split}
    \end{equation}
    
    And the FPR difference:
    
    \begin{equation}
    \begin{split}
        FPR_{dif} & = 
        |P(\hat{y}=1|y=0,g=0) \\
        &-P(\hat{y}=1|y=0,g=1)|
    \end{split}
    \end{equation}
    
    Returning the largest of both values: 

    \begin{equation}
        EOD_{dif} = max(TPR_{dif}, FPR_{dif})
    \end{equation}
    
    \item \textbf{Predictive Equality}~\cite{chouldechova2017fair} requires both protected and unprotected groups to have the same false positive rate (FPR), which is related to the fraction of subjects in the negative class that were incorrectly predicted to have a positive value. We obtain the Predictive Equality difference:

    \begin{equation}
    \begin{split}
        PE_{dif} & =
        |P(\hat{y}=1|y=0,g=0) \\
        & -P(\hat{y}=1|y=0,g=1)|
    \end{split}
    \end{equation}

    \item \textbf{Equal Opportunity}~\cite{chouldechova2017fair} is obtained if both protected and unprotected groups have an equal false negative rate (FNR), the probability of an individual from the positive class to have a negative predictive value. We use the Equal Opportunity difference:

    \begin{equation}
    \begin{split}
        EOP_{dif} & = 
        |P(\hat{y}=0|y=1,g=0) \\
        &-P(\hat{y}=0|y=1,g=1)|
    \end{split}
    \end{equation}
\end{itemize}

\section{Results}
\label{sec:results}

Our goal is to analyze the robustness of label correction methods in terms of predictive accuracy as well as fairness, considering models trained in 3 different ways -- $M_o$, $M_n$, $M_c$ --, first analyzing the similarity between the original labels and the ones obtained after applying each label noise correction method to the noisy data.

\subsection{Similarity to original labels after correction}

\begin{figure}[ht]
    \centering
    \includegraphics[width=0.45\textwidth]{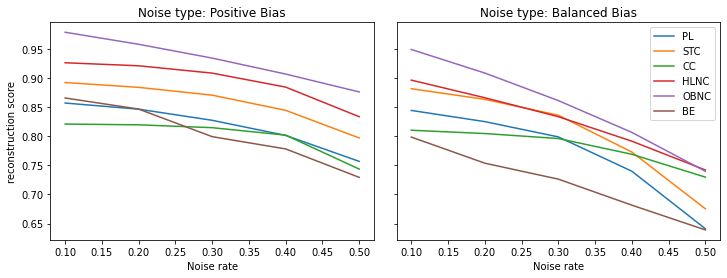}
    \caption{Reconstruction score ($r$), representing the similarity between original labels and the ones obtained after applying each label noise correction method for different noise rates.}
    \label{fig:similarity}
\end{figure}

Fig.~\ref{fig:similarity} shows, on average, how similar each method's correction was to the original labels, considering both types of bias. Regardless of the type of bias, OBNC was the method that was able to achieve higher similarity to the original labels. 

\subsection{Performance evaluation on the noisy test set}

\begin{figure}[!ht]
    \centering
    \includegraphics[width=0.47\textwidth]{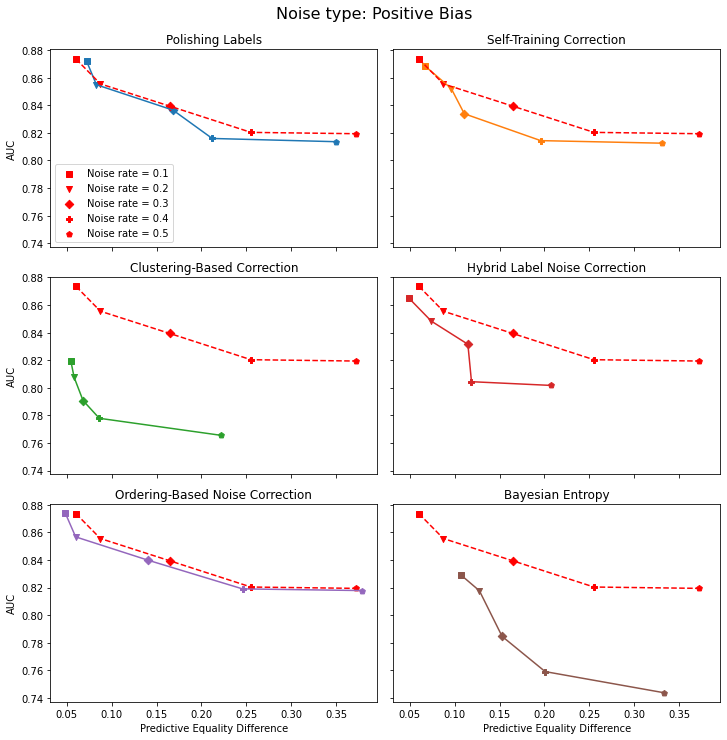}
    \caption{Trade-Off between AUC and Predictive Equality difference obtained on the \textit{noisy} test set when correcting the data injected with Positive Bias noise at different rates using each of the label correction methods. The red dashed line shows the performance of the model obtained from the \textit{noisy} train set at each noise rate.}
    \label{fig:noisy_test_positive_bias}
\end{figure}

\begin{figure}[!ht]
    \centering
    \includegraphics[width=0.47\textwidth]{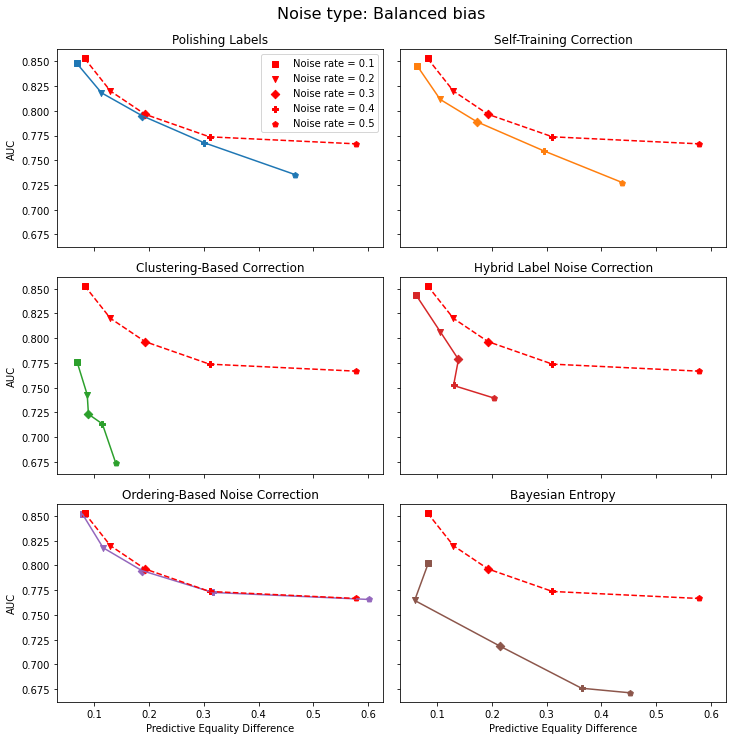}
    \caption{Trade-Off between AUC and Predictive Equality difference obtained on the \textit{noisy} test set when correcting the data injected with Balanced Bias noise at different rates using each of the label correction methods. The red dashed line shows the performance of the model obtained from the \textit{noisy} train set at each noise rate.}
    \label{fig:noisy_test_balanced_bias}
\end{figure}

In some cases, we may only have access to biased data both for training and testing the models. As such, we evaluate the predictive performance and fairness of the predictions of the models on the noisy test set. The trade-off between the AUC metric and the Predictive Equality difference metric for different noise rates is shown in Fig.~\ref{fig:noisy_test_positive_bias}, for the \textbf{Positive Bias} noise, and in Fig.~\ref{fig:noisy_test_balanced_bias}, for the \textbf{Balanced Bias} noise. In the remainder of this section, we only present the results in terms of Predictive Equality difference since the same conclusions can be derived from the results that were obtained using any of the aforementioned fairness metrics.

The OBNC method achieved performance similar to using the \textit{noisy} data, while PL and STC show small improvements, mainly in terms of fairness. The CC method performs the best in achieving fairness, being able to keep discrimination at a minimum even at higher noise rates, as shown in Figure~\ref{fig:noisy_test_balanced_bias}, but losing significant predictive performance to do so. The HLNC method maintained its ability to improve fairness at minimum expense to the predictive performance of the resulting models.

\subsection{Performance evaluation on the original test set}

\begin{figure}[!ht]
    \centering
    \includegraphics[width=0.47\textwidth]{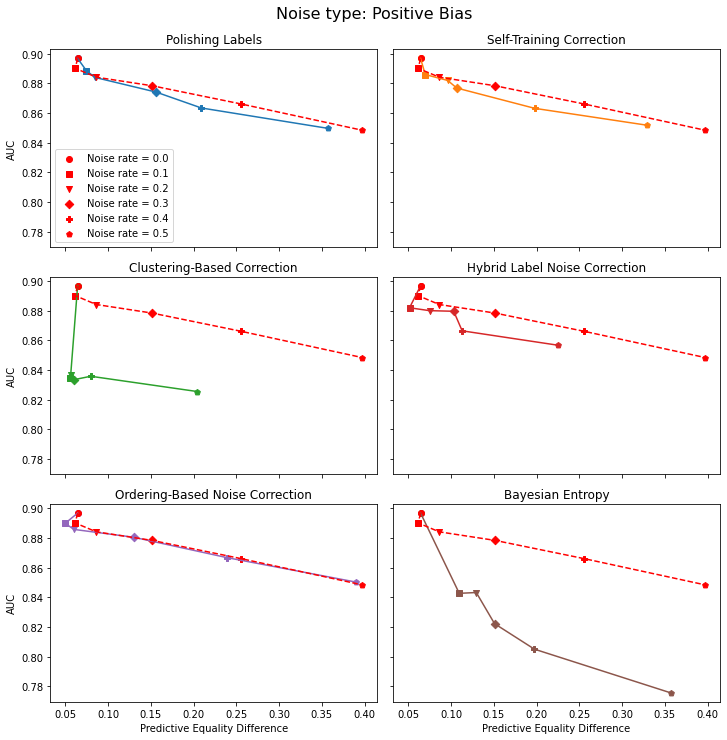}
    \caption{Trade-Off between AUC and Predictive Equality difference obtained on the \textit{original} test set when correcting the data injected with Positive Bias noise at different rates using each of the label correction methods. The red dashed line shows the performance of the model obtained from the \textit{noisy} train set at each noise rate.}
    \label{fig:original_test_positive_bias}
\end{figure}

\begin{figure}[!ht]
    \centering
    \includegraphics[width=0.47\textwidth]{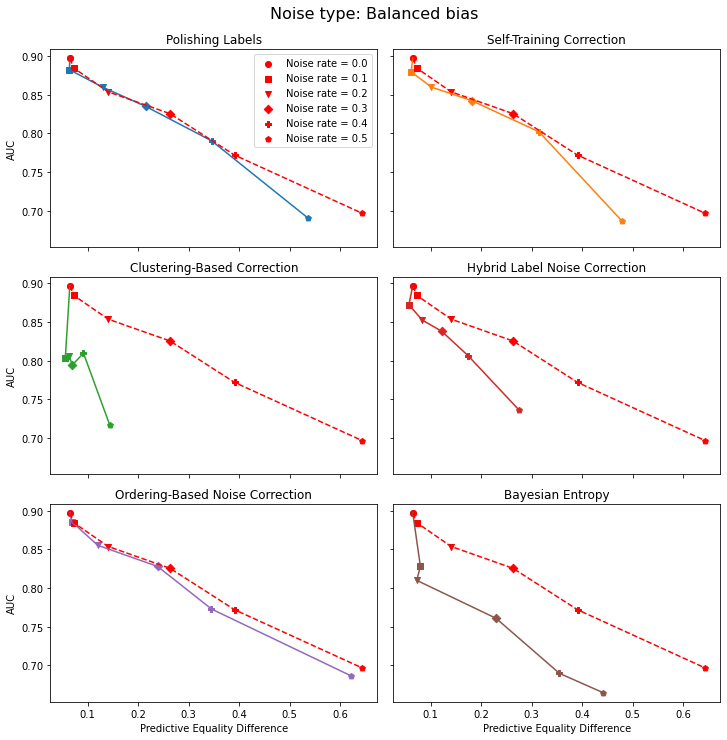}
    \caption{Trade-Off between AUC and Predictive Equality difference obtained on the \textit{original} test set when correcting the data injected with Balanced Bias noise at different rates using each of the label correction methods. The red dashed line shows the performance of the model obtained from the \textit{noisy} train set at each noise rate.}
    \label{fig:original_test_balanced_bias}
\end{figure}

To understand how the label noise correction methods fare on producing accurate and fair predictions from biased data in an environment where these biases are no longer present, we evaluate the performance of the three obtained models on the original test set. The trade-off between the AUC metric and the Predictive Equality difference for each method at different noise rates is shown in Fig.~\ref{fig:original_test_positive_bias}, for the \textbf{Positive Bias} noise, and in Fig.~\ref{fig:original_test_balanced_bias}, for the \textbf{Balanced Bias} noise.

In this testing scenario, the methods still behave in a similar way to the previous one in relation to each other. The OBNC method was shown to correct the labels in a way that is the most similar to the \textit{original} train set. Still, the performance of the resulting model is comparable to using the \textit{noisy} train set. The PL and STC methods achieve a slightly better trade-off between predictive performance and fairness. On the other hand, the CC method shows significant improvements in terms of fairness, but at the expense of a lower AUC score. The BE method achieves a low score in both metrics. Finally, the HLNC method was found to be the best at simultaneously improving both predictive performance and fairness.

\subsection{Performance evaluation on the corrected test set}

\begin{figure}[ht]
    \centering
    \includegraphics[width=0.47\textwidth]{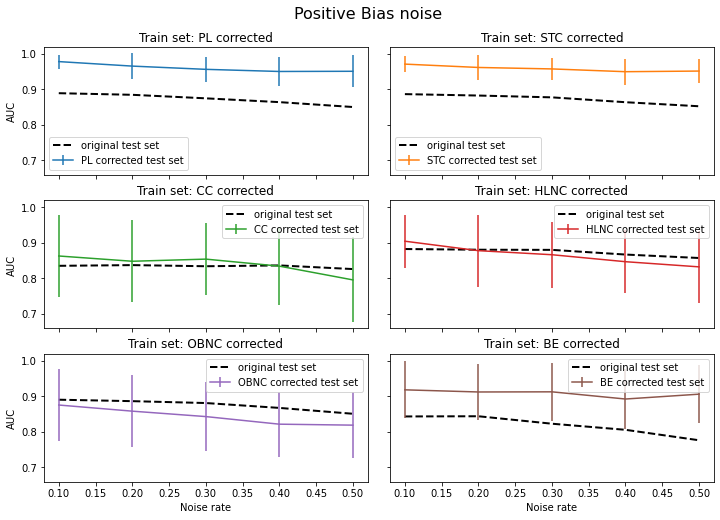}
    \caption{Comparison in AUC between testing the model obtained from the data corrected by each method on the original test set and on the test set corrected by the same method, in the presence of Positive Bias noise.}
    \label{fig:corrected_test_auc_positive_bias}
\end{figure}

\begin{figure}[ht]
    \centering
    \includegraphics[width=0.47\textwidth]{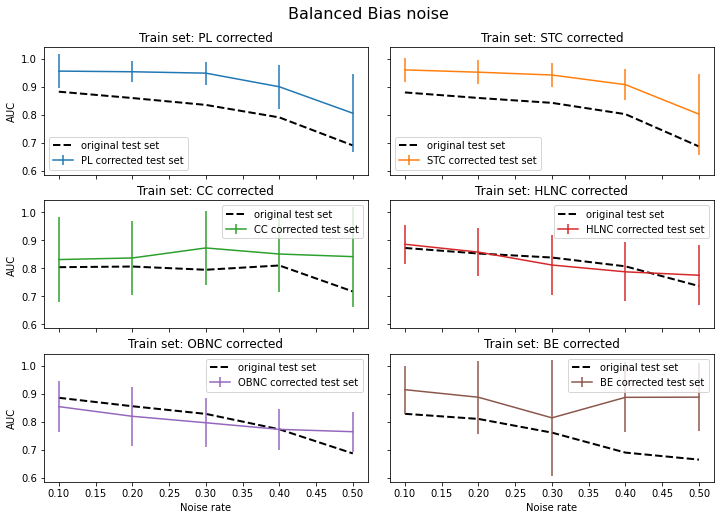}
    \caption{Comparison in AUC between testing the model obtained from the data corrected by each method on the original test set and on the test set corrected by the same method, in the presence of Balanced Bias noise.}
    \label{fig:corrected_test_auc_balanced_bias}
\end{figure}

\begin{figure}[ht]
    \centering
    \includegraphics[width=0.47\textwidth]{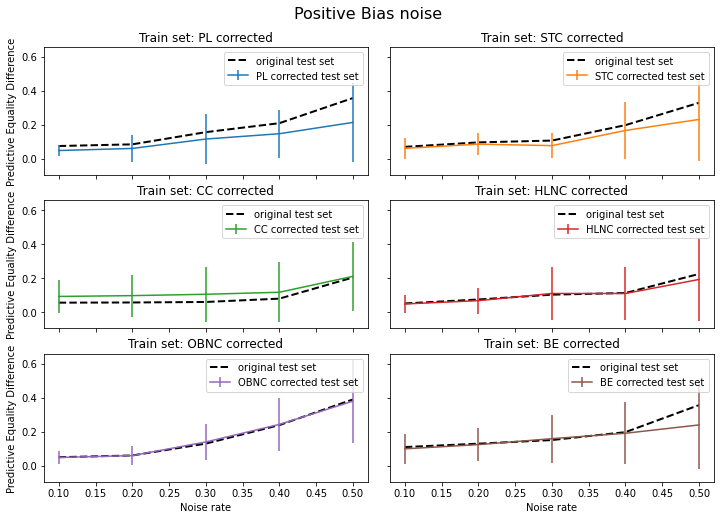}
    \caption{Comparison in Predictive Equality difference between testing the model obtained from the data corrected by each method on the original test set and on the test set corrected by the same method, in the presence of Positive Bias noise.}
    \label{fig:corrected_test_predequal_positive_bias}
\end{figure}

\begin{figure}[ht]
    \centering
    \includegraphics[width=0.47\textwidth]{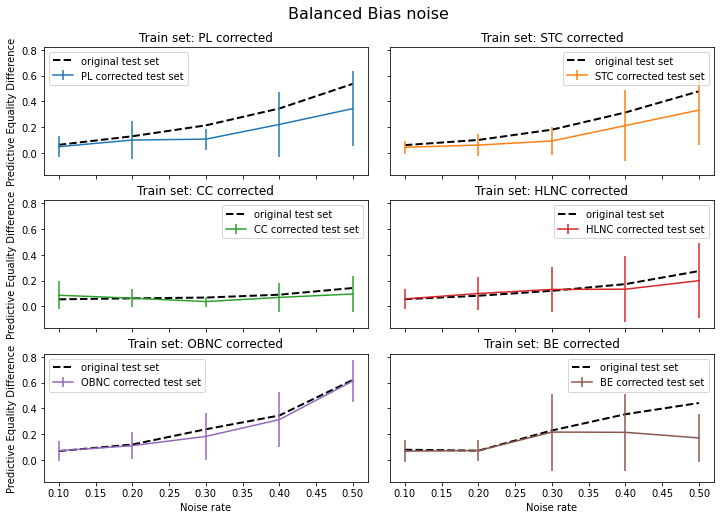}
    \caption{Comparison in Predictive Equality difference between testing the model obtained from the data corrected by each method on the original test set and on the test set corrected by the same method, in the presence of Balanced Bias noise.}
    \label{fig:corrected_test_predequal_balanced_bias}
\end{figure}

Finally, we investigate the possibility of applying label noise correction methods on the corrupted test set to simulate having an unbiased testing environment when only corrupted data is available for testing. To do so, we evaluate the performance of the models obtained using corrected train data on the test set corrected using the same method. We then assess whether that performance is similar to the one obtained when testing the same models on the original test set. The results for the AUC metric are presented in Fig.~\ref{fig:corrected_test_auc_positive_bias}, for the Positive Bias noise type, and in Fig.~\ref{fig:corrected_test_auc_balanced_bias}, for the Balanced Bias noise type. Considering the Predictive Equality difference metric, the results are shown in Fig.~\ref{fig:corrected_test_predequal_positive_bias} for the Positive Bias noise type, and in Fig.~\ref{fig:corrected_test_predequal_balanced_bias}, for the Balanced Bias noise type.

In terms of AUC, the PL, STC, and BE methods tend to result in an overestimation of the predictive performance of the resulting model. At the same time, the OBNC method appears to slightly underestimate it. The HLNC method shows similar performance to testing on the \textit{original} test set in the presence of both types of noise, while the CC method only achieves this when dealing with Positive Bias noise. Regarding the Predictive Equality difference metric, all methods show a performance very similar to using the \textit{original} test set. A slight underestimation of discrimination can be seen for the PL, STC, and BE methods for the higher noise rates.

\section{Discussion}
\label{sec:discussion}

The ability to correct the labels does not necessarily guarantee a good compromise between accuracy and fairness. For instance, the OBNC method obtained the highest similarity with the original labels. However, when assessing the compromise between predictive performance and fairness, the OBNC method had a much less satisfactory performance, showing barely any difference from training with the noisy training set. 

On the other hand, the CC method, which did not show a high reconstruction score, kept discrimination at minimum values, even at the highest noise rates. However, this was achieved at the cost of lower predictive performance. The nature of the fairness metrics can explain this: e.g., the Predictive Equality metric calculates the difference between the FPR of each group, meaning that if both groups have a high but similar FPR, the predictions are technically fair but not accurate. 

We must acknowledge some limitations of this study, as they can impact the generalizability of our findings. 
The first one is related to an important advantage of the proposed methodology: it may use standard benchmark datasets to assess the robustness of label correction methods. This means that analysis can be based on a much larger set of datasets than typical fairness studies use. However, the choice of both the sensitive attribute and the positive class is arbitrary. This means that these datasets do not necessarily have similar distributions to real problems with label noise caused by discrimination. However, the methodology can also be applied to benchmark fairness datasets to assess the generality of the results obtained. 
Additionally, the predicted classes were based on a threshold of 0.5, which is not realistic in many problems where discrimination might be an issue. As the choice of threshold impacts the fairness metrics, it is important to obtain results with other thresholds. In the case of benchmark fairness datasets, problem-specific thresholds can also be used.
Therefore, future research would involve applying this methodology to benchmark fairness datasets.

\section{Conclusions}
\label{sec:conclusions}

In this work, we tackle the problem of learning fair ML classifiers from biased data. In such a scenario, we look at the inherent discrimination in datasets as label noise that can be eliminated using label noise correction techniques. This way, the corrected data could be used to train fair classifiers using standard ML algorithms without further application of fairness-enhancing techniques. We propose a methodology to empirically evaluate the effect of different label noise correction techniques in improving the fairness and predictive performance of models trained on previously biased data. Our framework involves manipulating the amount and type of label noise and can be used on both fairness benchmarks and standard ML datasets. In the conducted experiments, we analyzed six label noise correction methods. We observed that the Hybrid Label Noise Correction method was able to achieve the best trade-off between fairness and predictive performance.

\bibliographystyle{IEEEtran}
\bibliography{main}
\end{document}